\documentclass[10pt,twocolumn,letterpaper]{article}

\usepackage{cvpr}              

\usepackage{graphicx}
\usepackage{amsmath}
\usepackage{amssymb}
\usepackage{booktabs}
\usepackage{algorithm}
\usepackage[accsupp]{axessibility}  
\usepackage[pagebackref,breaklinks,colorlinks]{hyperref}

\usepackage{amsmath,amsfonts,bm}

\def\eqref#1{equation~\ref{#1}}

\def\1{\bm{1}}

\DeclareMathAlphabet{\mathsfit}{\encodingdefault}{\sfdefault}{m}{sl}
\SetMathAlphabet{\mathsfit}{bold}{\encodingdefault}{\sfdefault}{bx}{n}

\usepackage{amsmath}
\usepackage{mathtools}

\usepackage[capitalize]{cleveref}
\crefname{section}{Sec.}{Secs.}
\Crefname{section}{Section}{Sections}
\Crefname{table}{Table}{Tables}
\crefname{table}{Tab.}{Tabs.}

\def\cvprPaperID{11131} 
\def\confName{CVPR}
\def\confYear{2023}

\begin{document}


\title{You Need Multiple Exiting: Dynamic Early Exiting for \\ Accelerating Unified Vision Language Model}


\author{
Shengkun Tang$^1$, Yaqing Wang$^2$, Zhenglun Kong$^6$, Tianchi Zhang$^4$, Yao Li$^5$, Caiwen Ding$^3$\\
Yanzhi Wang$^6$, Yi Liang$^2$, Dongkuan Xu$^1$\thanks{Corresponding author}\\
$^1$North Carolina State University, Raleigh, USA \quad $^2$Google Research, New York, USA\\
 $^3$University of Connecticut, Mansfield, USA\quad$^4$University of Michigan, Ann Arbor, USA\\$^5$The University of North Carolina at Chapel Hill, Chapel Hill, USA \quad $^6$Northeastern University, Boston, USA  \\
\tt\small shengkuntangwork@gmail.com, \{yaqingwang, yiliang\}@google.com, yaoli@email.unc.edu\\ \tt\small\{kong.zhe, yanz.wang\}@northeastern.edu, tonyztc@umich.edu,  caiwen.ding@uconn.edu, dxu27@ncsu.edu
}\maketitle

\begin{abstract}

{Large-scale Transformer models bring significant improvements for various downstream vision language tasks with a unified architecture. The performance improvements come with increasing model size, resulting in slow inference speed and increased cost for severing. While some certain predictions benefit from the full computation of the large-scale model, not all of inputs need the same amount of computation to conduct, potentially leading to computation resource waste. To handle this challenge, early exiting is proposed to adaptively allocate computational power in term of input complexity to improve inference efficiency. The existing early exiting strategies usually adopt output confidence based on intermediate layers as a proxy of input complexity to incur the decision of skipping following layers. However, such strategies cannot be applied to encoder in the widely-used unified architecture with both encoder and decoder due to difficulty of output confidence estimation in the encoder layers. It is suboptimal in term of saving computation power to ignore the early exiting in encoder component. To address this issue, we propose a novel early exiting strategy for unified vision language models, which allows to dynamically skip the layers in encoder and decoder simultaneously in term of input layer-wise similarities with multiple times of early exiting, namely \textbf{MuE}. By decomposing the image and text modalities in the encoder, MuE is flexible and can skip different layers in term of modalities, advancing the inference efficiency while minimizing performance drop. Experiments on the SNLI-VE and MS COCO datasets show that the proposed approach MuE can reduce expected inference time by up to 50\% and 40\% while maintaining 99\% and 96\% performance respectively. Our source code has already been merged into  official OFA repository and is available at \url{https://github.com/OFA-Sys/OFA}.
}
\end{abstract}
\section{Introduction}
\label{sec:intro}


Recent advances in multi-modal Transformer-based large-scale models~\cite{li2020oscar,lin2021m6,wang2021simvlm,zhang2021vinvl} bring improvements across various vision language tasks. Among the Transformer-based models, the unified sequence-to-sequence architecture~\cite{wang2022ofa,  2020t5} has attracted much attention due to its potential to become an universal computation engine to diverse tasks. Although large-scale models have achieved unattainable performance, their expensive computational cost hinders their applications in real-time scenarios.

\begin{figure}[t]
\begin{center}
\includegraphics[width=0.4\textwidth]{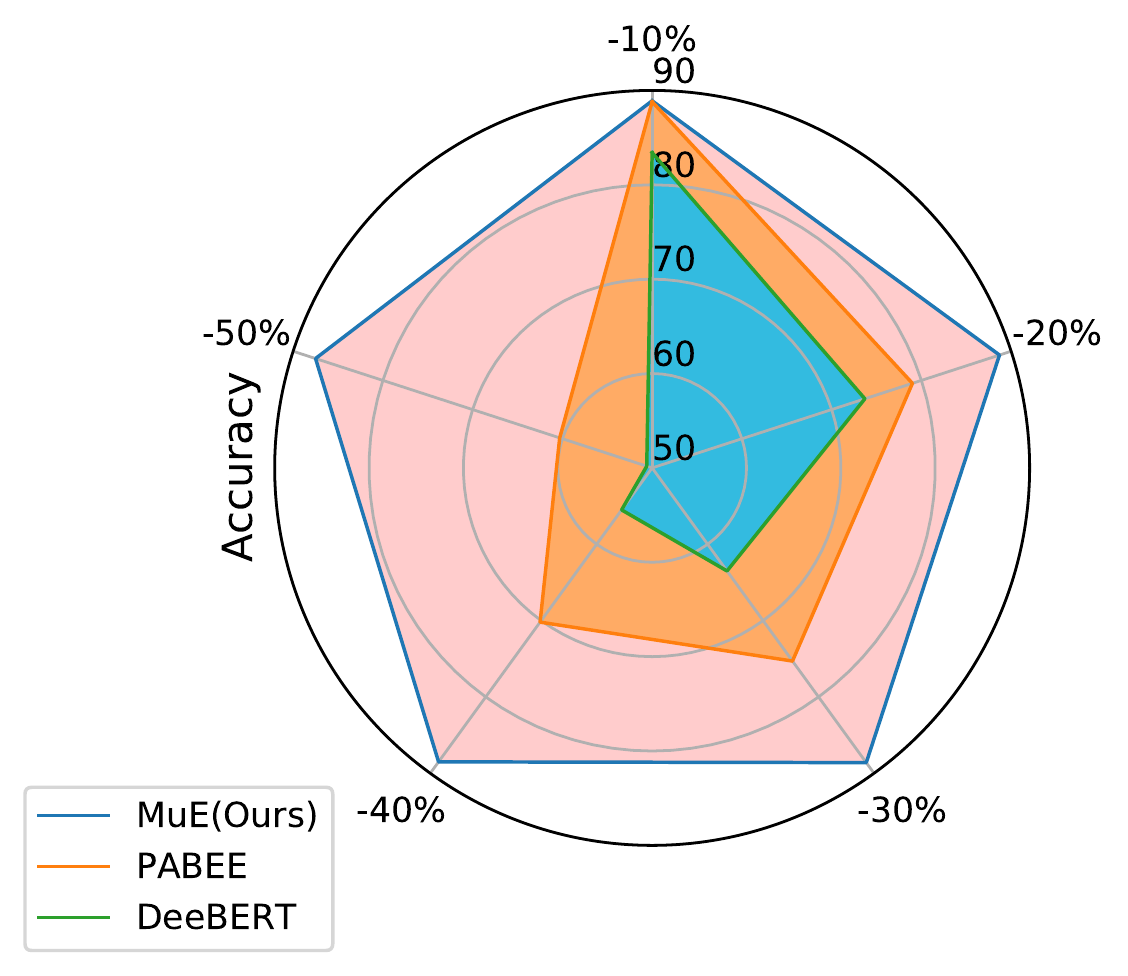}
\includegraphics[width=0.4\textwidth]{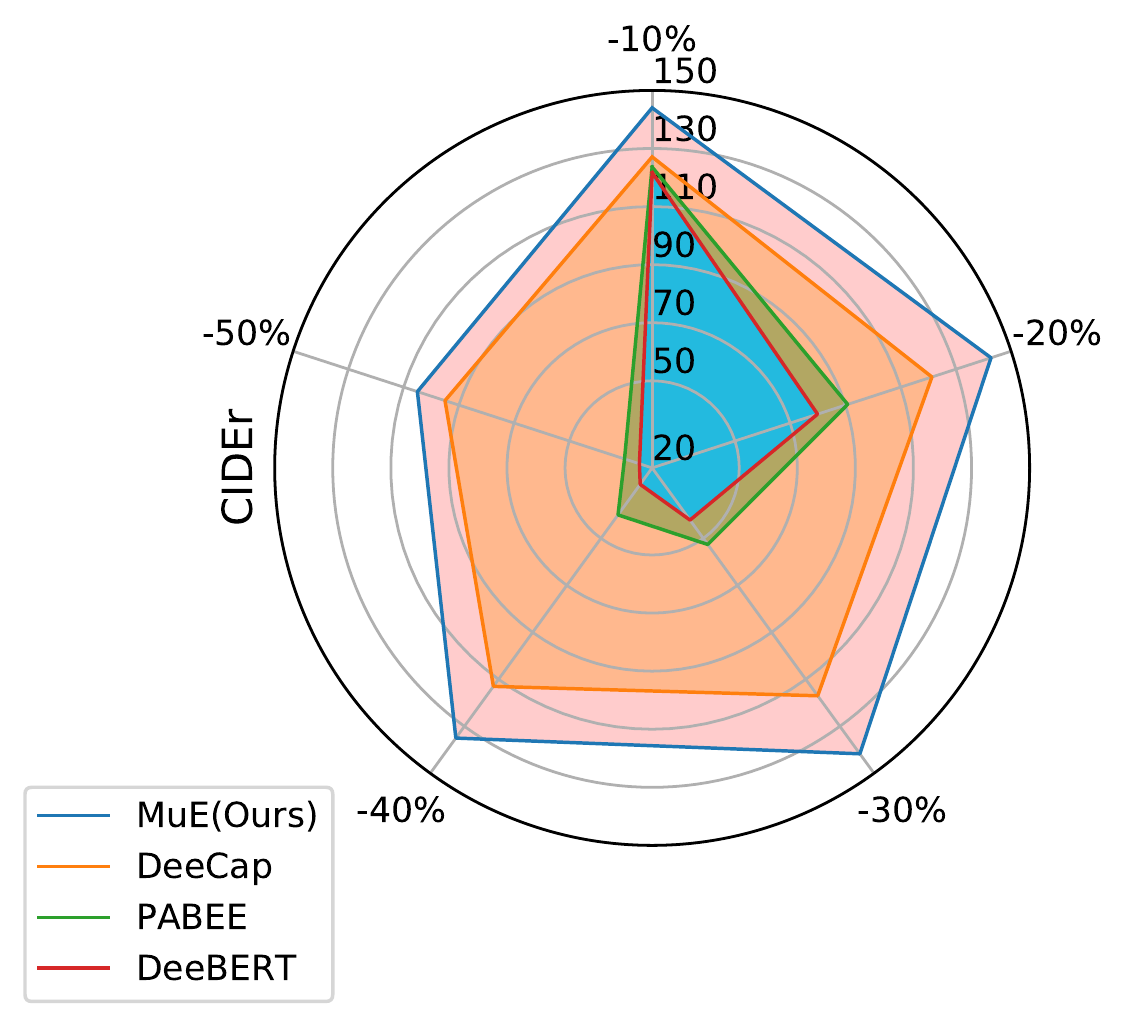}
\end{center}
\caption{The performance of different early exiting methods on SNLI-VE~\cite{xie2019visual} and MS COCO~\cite{chen2015microsoft} with certain expected inference time reduction rates.}
\label{Fig:inspiration}
\end{figure}

While the scaling effect suggests that the performance of the model benefits from its increased size, not every input requires the same amount of computation to achieve similar performance. Such an observation is particularly valid in visual language tasks, where inputs from different modalities may require different amounts of computation. Early exiting is a popular method to reduce computational costs by adaptively skipping layers on top of the input while preserving the general knowledge of the large-scale model.
Existing studies aim to deal with early exiting  for encoder-only models~\cite{xin2021berxit, xin2020deebert} or decoder components in encoder-decoder architectures~\cite{fei2022deecap}, but cannot induce early exiting decisions for both components at the same time. Considering that single-component strategies may be suboptimal in terms of saving computation cost, in this paper, we investigate how to perform early exiting for both encoder and decoder components in a sequence-to-sequence architecture to elucidate a new way to further improve inference efficiency.

Given the varied complexity of the inputs, it is natural to consider skipping some layers of the encoder as well as the decoder. Current decision mechanisms use classifiers to predict the output confidence of intermediate representations and stop computation if the confidence reaches predefined threshold. However, extending this to unified sequence-to-sequence model is non-trivial due to two main challenges: (1) there are dependencies in making decisions for exiting decisions in the encoder and decoder, and (2) it is difficult to apply confidence classifiers to skip the encoder layer before going through the decoder layer for task output.
To address the above challenges and to enable early exiting of encoder and decoder in sequence-to-sequence framework, we propose a novel early exiting strategies based on layer-wise input similarity, which is different from existing works based on task confidence~\cite{schuster2021consistent}. More specifically, the model is encouraged to skip following layers in both encoder and decoder when the layer-wise similarity reaches a certain threshold. 

This method is inspired by the saturation observation~\cite{geva2022transformer} which shows that the hidden-state of each Transformer layer arrives at a saturation status as going into deep layers. For the vision-language tasks, we find that the similar observation regarding saturation is also valid as shown in Figure~\ref{Fig:saturation}. 
This observation lands the foundation that we could make the exiting decision based on the intermediate layer-wise input similarities without going through the following layers. Besides, since the computation needed for input in different modalities usually varies, we propose a modality decomposition mechanism, which could further enable early fusion large-scale multi-modal models to break tie between modalities and enjoy the flexible exiting decision over modalities. To encourage the early exiting behavior with a minimal performance loss, we design a layer-wise task loss, which enforce the each layer to output informative features for final task. Figure~\ref{Fig:inspiration} shows the results on SNLI-VE dataset\cite{xie2019visual} and MS COCO\cite{chen2015microsoft} in term of expected time reduction rate and task performance. We compare MuE with several State-of-the-art early existing methods and observe that MuE is able to largely reduce inference time with a minimal performance drop compared to other SoTA methods. 
Our main contributions are summarized as follows:
\begin{itemize}
    \item To the best of our knowledge, this is a pioneering work to extend early exiting choices to both encoder and decoder of sequence-to-sequence architecture. To this end, we propose a novel early exiting strategy based on layer-wise input similarity with the valid assumption on saturation states in vision language models. 
    
    \item Given the different characteristics of the modalities and tasks, we decompose the modalities encoder for early-fusion pre-trained models, bringing additional improvements in terms of inference efficiency.
    
    \item We introduce layer-wise task loss, linking each layer in the decoder to the final task, effectively helping to maintain task performance when a significant time reduction is required.
    
    \item Extensive experiments show that our method can largely reduce the inference time of vision language models by up to 50\% with minimal performance drop.
\end{itemize}

\section{Related Work}
\label{sec:relatedwork}
\noindent {\bf Vision Language.} Vision language learning recently attracts lots of attention~\cite{zhou2020unified,tan2019lxmert, li2020unicoder, chen2020uniter, lin2020interbert,lu202012mt, xu2021e2e, li2020oscar, zhang2021vinvl, gan2020large, dou2022empirical, radford2021learning, li2022grounded, zhong2022regionclip, yang2022unifiedlabel, wang2021vlmo}, where Transformer-based model~\cite{li2019visualbert, lu2019vilbert} brings significant improvements to diverse downstream vision language tasks. Recent efforts~\cite{li2020oscar,lin2021m6,wang2021simvlm,zhang2021vinvl, yang2021crossing, yu2022coca}  leverage the sequence-to-sequence architecture to unify diverse tasks in a generation manner to build the universal computation engine. With the unified framework, the recent study \cite{wang2022ofa} shows promising results on several vision language benchmark datasets via introducing instructions to handle diverse tasks. Even though the large-scale vision language models achieve unattainable performance, the expensive inference costs still hinder their application in real-world scenarios.

\begin{figure*}[t]
\begin{center}
\includegraphics[width=0.85\textwidth]{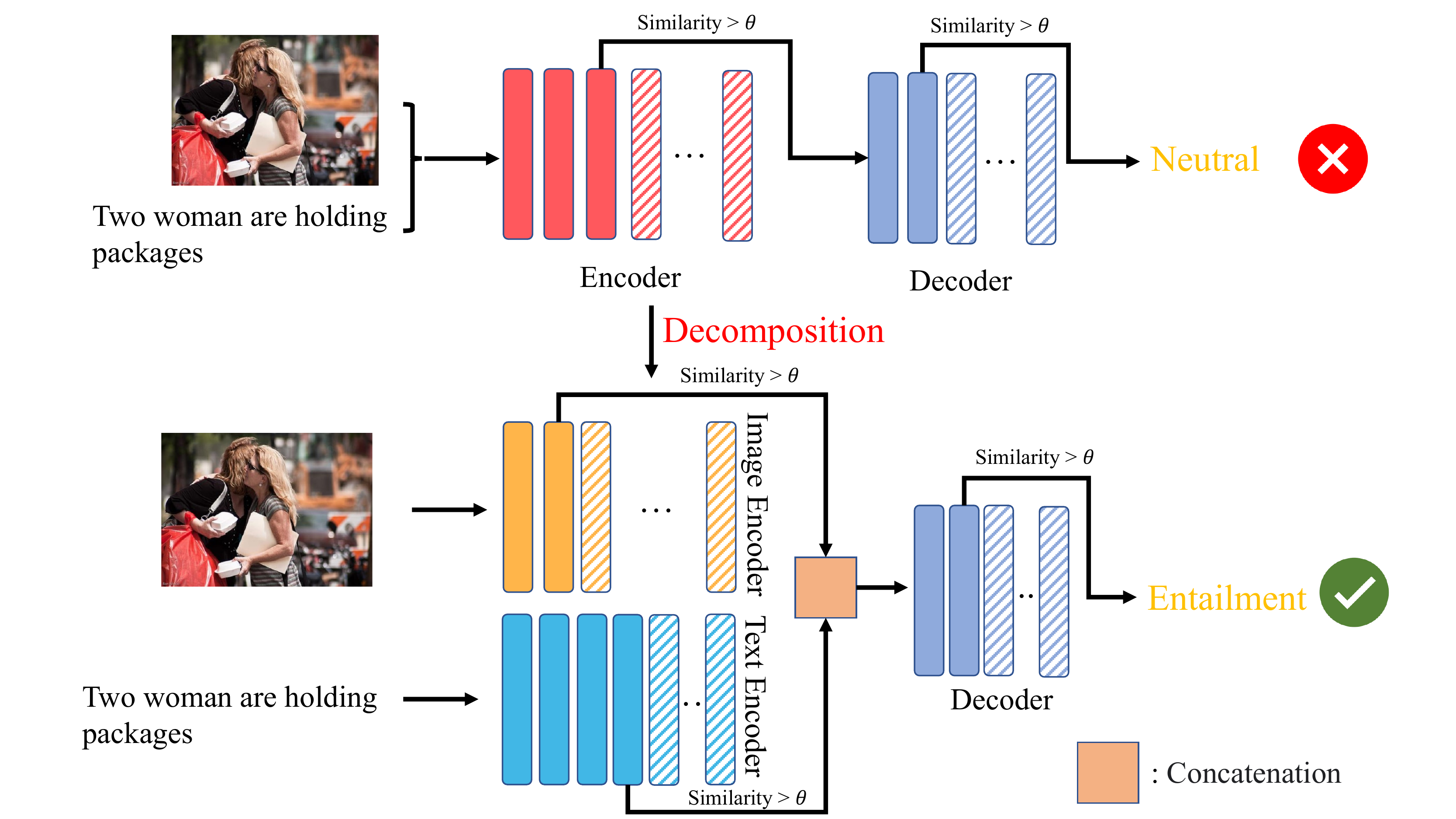}
\end{center}
\caption{The overview of our proposed MuE.}
\label{Fig:Architecture}
\end{figure*}

\noindent {\bf Early Exiting Strategy.} How to improve inference efficiency of large-scale models attracts lots of attention in recent years with existing efforts including knowledge distillation \cite{hinton2015distilling, fang2021compressing}, quantization\cite{liu2021post}, weight pruning \cite{gan2022playing} and others \cite{kong2021spvit, rao2021dynamicvit, li2017not, liang2022notall}. In this paper, we focus on early exiting~\cite{huang2017multi, li2019improved, li2017not, phuong2019distillation, panda2016conditional,teerapittayanon2016branchynet, kaya2019shallow,xin2021berxit, elbayad2019depth, liu2021faster}, which aims to dynamically allocate computation resource per example, with the goal of improving the inference speed while minimizing the performance drop. DeeBERT~\cite{xin2020deebert} and FastBERT~\cite{liu2020fastbert} explore early exiting strategy with BERT~\cite{devlin2018bert} model and several efforts are on varied natural language processing tasks including natural language understanding~\cite{liao2021global,li2020cascadebert,zhou2020bpatience}, sequence labeling~\cite{li2021accelerating} and document ranking\cite{xin2020early}. For the vision language learning, DeeCap~\cite{fei2022deecap} explores early exiting in image captioning tasks  with a shallow imitation network to help maintain the performance. However, DeeCap is dedicated to decoder layers only and cannot be easily extended to encoders, thereby limiting the room of further advancing the inference efficiency. To handle this challenge, we propose a novel early exiting strategy, which introduces this computation resource allocation process in both encoder and decoder, while minimizing the performance drop.

\section{Method}

The objective of early exiting is to dynamically allocate the computation power in term of input complexity to improve the inference efficiency while maintaining the performance. Considering a unified visual language model consisting of encoder and decoder, we propose an early exiting method which is able to  skip layers in both encoder and decoder components.  The input to the visual language model includes images and text, while the output varies with the need of downstream tasks, such as the textual sentence for image captioning and class prediction for visual entailment.

The overview of the proposed approach is given in Figure~\ref{Fig:Architecture}. To allow flexibility in the decision to skip different modalities, we discuss how to decompose the modalities in Sec.~\ref{sec:decomposition}. Then, we describe how to use layer-wise similarity to guide early exiting of encoder and decoder layers in Sec.~\ref{sec:earlyexit}. In Sec.~\ref{sec:training}, we introduce layer-wise task loss, which can be effective in helping to maintain performance when a significant reduction in inference cost is required.


\subsection{Modality Decomposition}
\label{sec:decomposition}
{We introduce how to decompose the early-fusion encoder to modality-specific encoder to allow processing image and text independently during fine-tuning and inference stages. We first describe how the image and text are processed in the early-fusion encoder of unified vision language architecture. Following the recent work~\cite{wang2022ofa}, we represent text and image in tokens. Let the input token embedding representations of image be $\mathbf{I_\mathrm{0}}$ and text be $\mathbf{T_\mathrm{0}}$. Before feeding into transformer layers, the image tokens are processed by ResNet~\cite{he2016deep}. As for positional information, we decouple the position correlation from image embeddings and text embeddings~\cite{ke2020rethinking}. Moreover, we use 2D relative position bias\cite{wang2021simvlm, dai2021coatnet} for image and 1D relative position bias for text \cite{raffel2020exploring}, respectively. The full input sequence includes the concatenation of token representations from image and text tokens as $\mathbf{[I_\mathrm{0}; T_\mathrm{0}]}$. The Transformer encoder has $n$ layers (denoted $E_i$ for layer $i$) and we refer the readers to \cite{vaswani2017attention} for the details of the Transformer layer. Thus, the process can be formulated as:  }
\begin{equation}
    \mathbf{[I_\mathit{i}; T_\mathit{i}]} = E_i(\mathbf{[I_{\mathit{i}-\mathrm{1}}; T_{\mathit{i}-\mathrm{1}}]}).
\end{equation}
 We denote a stack of layers from layer $i$ to layer $j$ in encoder as $E_{i:j}$. The image $\mathbf{I_\mathit{n}}$ and text input $\mathbf{T_\mathit{n}}$ through the encoder without decomposition strategy can be written as:
\begin{equation}
    \mathbf{[I_\mathit{n}; T_\mathit{n}]} = E_{1:n}\mathbf{([I_\mathrm{0}; T_\mathrm{0}])}.
\end{equation}

{To decompose the encoder into modality-specific encoder, we duplicate the encoder to handle the input, where the image tokens and text tokens are fed into two encoder respectively. The parameters of the two encoder are tied without introducing extra parameters. The output representations of the decomposed encoder, $\mathbf{I_\mathit{n}}$ and $\mathbf{T_\mathit{n}}$ can be expressed as:}
\begin{equation}
    \mathbf{[I_\mathit{n}; T_\mathit{n}]} = [E_{1:n}(\mathbf{I_\mathrm{0}}); E_{1:n}(\mathbf{T_\mathrm{0}})].
\end{equation}


Encoder decomposition enables the proposed model to process the image and text independently, which in turn allows us to apply early exiting for different modalities by necessity and pick up essential information of each modality for different tasks during inference. When the image encoder exits at layer $p$ and the text encoder exits at layer $q$, The output of the encoder part can be written as:
\begin{equation}
    [\mathbf{I_\mathit{p}}; \mathbf{T_\mathit{q}}] = [E_{1:p}(\mathbf{I_\mathrm{0}});E_{1:q}(\mathbf{T_\mathrm{0}})].
\end{equation}

After decomposition, the image and text tokens are concatenated and fed into the decoder, denoted as $C=\mathbf{[I_p; T_q]}$. The decoder contains a self-attention module and a cross-attention module in each layer. We denote the input of self-attention module as $Td_{i, s}, i=0$. $s$ refers to the step in decoding. 
In the layer $i$, The output representation of self-attention module and $C$ will be sent into cross-attention module, which is formulated as following:
\begin{align}
     \textit{Td}_{i, s} &= \text{CrossAttn}(\text{SelfAttn}(\textit{Td}_{i-1, s}), C),\\
    \textit{Td}_{0,s} &= \textit{Td}_{n,s-1}.
\end{align}

The final output is decoded via a unified vocabulary following \cite{wang2022ofa} to obtain the textual output.

\subsection{Early Exiting Based on Layer-wise Similarities}
\label{sec:earlyexit}
Existing early exiting methods\cite{fei2022deecap, xin2020deebert, schuster2021consistent} usually add a classifier for each layer to simulate output for input complexity estimation. During inference, the output representations of each layer are fed into the classifier and the confidence or entropy are utilized as a proxy for early exiting decision. However, the classifier cannot be applied to the encoder part in encoder-decoder framework since the hidden-state representation of encoder is not directly related to the final prediction considering that the hidden-state representation will still go through the following decoder.
According to \cite{geva2022transformer}, the hidden-state representation of each layer in Transformer will reach a saturation status in language model, which indicates that the hidden-state representation change decreases as going through the latter layers. Therefore, the latter layers can be skipped safely without significant performance drop when such saturation status is reached. 

\begin{figure}[t]
\begin{center}
\centering
\includegraphics[width=0.4\textwidth]{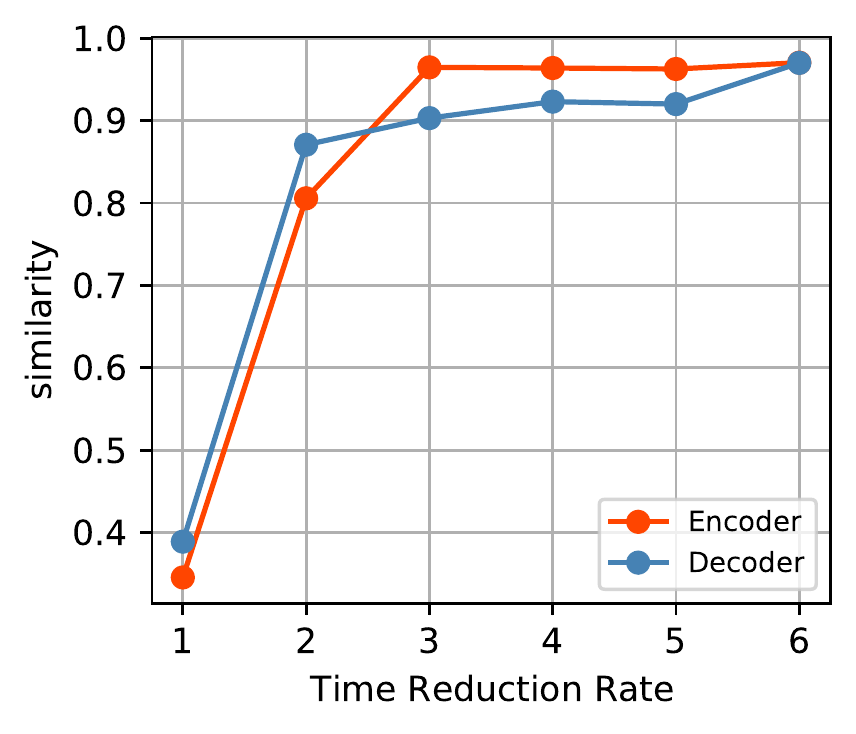}
\end{center}
\vspace{-2em}
\caption{The saturation status in vision language model. The similarities between hidden-states reach the top at an early stage and stay stable in the following layers.}
\label{Fig:saturation}
\end{figure}

We study whether the saturation state exists in multi-modal sequence-to-sequence models, as shown in Figure~\ref{Fig:saturation}, where the Cosine Similarity of layers is evaluated on the test set of SNLI-VE dataset~\cite{xie2019visual}. As observed in Figure~\ref{Fig:saturation}, the shallow layers of encoder and decoder show the low similarity level, indicating that the token representation changes dramatically in shallow layers. The similarities reach peak and remain flat in the following layers, confirming the existence of saturation. Based on this observation,  we leverage the Cosine Similarity between layers as a proxy to estimate the saturation level:
\begin{equation}
    \text{CosSim}(T_i, T_j) = \frac{T_i \cdot T_j}{\Vert T_i \Vert \cdot \Vert T_j \Vert}.
\end{equation}
We denote each layer in encoder and decoder as $E_i$ and $D_i(i<n)$. Image token, text token and decoder token representation are $I_i$, $T_i$, $Td_{i,s}$, respectively. The similarity can be formulated as:
\begin{align}
    \text{ImgSim}_i &= \text{CosSim}(E_i(\mathbf{I_\mathit{i}}), E_{i-1}(\mathbf{I_{\mathit{i}-\mathrm{1}}})),\\
    \text{TxtSim}_i &= \text{CosSim}(E_i(\mathbf{T_\mathit{i}}), E_{i-1}(\mathbf{T_{\mathit{i}-\mathrm{1}}})),\\
    \text{DecSim}_{i,s} &= \text{CosSim}(D_i(Td_{i,s}), D_{i-1}(Td_{i-1,s})).
\end{align}


Note that we compute the similarity between output representation of first layer and the input token representation when $i=1$. In that case, we can evaluate the saturation state of first layer. Once the similarity meets the predefined threshold, the computation will terminate and skip the following layers. During the inference stage in generation tasks such as image captioning, we generate new words in a auto-regressive manner and the proposed early exiting strategy apply to each step of generation. The detailed pseudo-codes are shown in Algorithm~\ref{inference}.

During the inference of generation tasks, the errors of each step will accumulate, which degrades the final generation results severely. We reduce errors at the beginning of generations with slight computation increase since the token numbers at early generation stage are relatively small. Therefore, we follow \cite{schuster2022confident} to utilize decay threshold $\Theta$ which can be formulated as follows:
\begin{equation}
    \Theta(t) = \beta \theta + (1-\beta)e^{-\tau t/N},
\end{equation}
where $\theta$ is a pre-define static threshold, $t$ is the timestep of generation, $N$ is the total number of steps, $\beta$ and $\tau$ are all hyperparameters.

\begin{algorithm}[t]
 \textbf{Input} image $I$, text tokens $T$, decoder input $T_{d,s}$, exiting threshold $\theta$
  \caption{Inference}\label{inference}
  \begin{algorithmic}[1]
   \STATE  $I = ResNet(I)$;  $T = Embed(T)$
   \STATE  $img\_states = [I]$; $txt\_states =[T]$
\vspace{4pt}
\item[] \text{\# For text encoder.}
    \FOR{$i \gets 1$ to $N$} 
    \STATE $T = \textit{$Enc_i$}(T)$
    \IF{$similarity(txt\_states[-1],T) > \theta$ }
     \STATE \textbf{break}
    \ENDIF
    \STATE $txt\_states.append[T]$
    \ENDFOR
\vspace{4pt}
\item[] \text{\# For image encoder.}
    \FOR{$i \gets 1$ to $N$}
    \STATE $I = Enc_i(I)$
    \IF{$similarity(img\_states[-1],I) > \theta$ }
     \STATE \textbf{break}
     \ENDIF 
     \STATE $img\_states.append[I]$
    \ENDFOR
\vspace{4pt}
\item[] \text{\# For Decoder.}
     \STATE $Enc\_out$ = $Concat[I,T]$; $Dec\_states$ = [$T_{d,0}$]
     \FOR{$s \gets 1$ to $S$}
    \FOR{$i \gets 1$ to $N$}
    \STATE $T_{d, s}$ = \textit{$Dec_i$}($T_{d,s}$; Enc\_out)
 \IF{$similarity(Dec\_states[-1], T_{d,s}) > \theta$}
     \STATE \textbf{break}
     \ENDIF
     \STATE $Dec\_states.append$[$T_{d,s}$]
     \ENDFOR
     \ENDFOR
    \STATE \textbf{Return}: $T_{d,s}$
\end{algorithmic}
\end{algorithm}

\begin{table*}[]
    \centering
    \begin{tabular*}{13.3cm}{l|ccc|ccccc}
    \hline
    \hline
        \multirow{2}*{Models} & \multicolumn{3}{c|}{SNLI-VE} & \multicolumn{5}{c}{Image Captioning} \\
        & Dev & Test & Time & BLEU-4 & METEOR & CIDEr & SPICE & Time \\
    \hline
       $\text{OFA}_{\textit{Base}}$ & 89.3 & 89.2 & 1 & 42.8 & 31.7 & 146.7 & 25.8 & 1 \\
    \hline
       $\text{OFA}_{\textit{Tiny}}$ & 85.3 & 85.2 & -33\% & 38.1 & 29.2 & 128.7 & 23.1 & -33\% \\
       DeeBERT & 78.9 & 78.8 & -15\% & 30.1 & 26.3 & 102.1 & 20.5 & -15.5\% \\
       PABEE & 85.3 & 85.2 & -15.3\% & 31.4 & 26.8 & 105.8 & 21 & -16.3\% \\
       DeeCap & - &  - &  - & 38.7 & 29.1 & 129 & 22.5 & -38\% \\
    \hline
       Ours & \textbf{88.7} & \textbf{88.5} & \textbf{-50\%} & \textbf{41.6} & \textbf{30.6} & \textbf{137} & \textbf{24.4} & \textbf{-40.2\%} \\
    \hline
    \hline
    \end{tabular*}
    \caption{The performance and expected time reduction rate comparison between our method and previous methods. Our method reduces more computation comparing with other methods while preserves well performance. Time: expected time reduction rate}
    \label{tab:result}
\end{table*}

\subsection{Layer-wise Task Loss}
\label{sec:training}
The main assumption of early exiting method is that the intermediate representations of some easy samples in test sets have enough information for final predictions~\cite{teerapittayanon2016branchynet, xin2020deebert}. However, as we show in Sec.~\ref{section:evalcaption}, the intermediate hidden states fail to predict the final results in decoding stage of sequence-to-sequence framework. Correspondingly, a significant performance drop is observed with intermediate representation of fine-tuned model. To address this issue, we propose to update every decoder layer of our model simultaneously with task loss to encourage the early exiting behavior and achieve the goal of maintaining the performance.

{The existing early exiting methods usually adopt two stage of training, where first one is to fine-tune the model and the following one is to develop an exiting classifier with the frozen fine-tuned checkpoint from the first step.  Different from the existing early exiting works, we introduce layer-wise task loss to encourage early exiting behavior during fine-tuning. More specifically, we add loss to each decoder layer and update them via optimizing the task loss. We observe that such a step could effectively improve the efficiency of early exiting and help maintain task performance. The loss function is written as:}
\begin{align}
      \mathcal{L} &= \frac{1}{N}\sum_i^N   \mathcal{L}_{\textit{CE}},\\
      \mathcal{L}_{\textit{CE}} &= -\sum_{i=1}^{|y|}\log P_\theta(y_i|y_{<i}, \mathbf{I}, \mathbf{T}),
\end{align}
where $N$ is the number of decoder layers, $\mathbf{I}$, $\mathbf{T}$ and $y$ denote image tokens, text tokens and output respectively. With this new training objective, the model can be optimized to balance severing final task from intermediate layer and final layer.  Therefore,  the prediction based on the intermediate is more likely to be informative, reducing the performance gap between  early exiting and full model. 

\section{Experiments}

\subsection{Experimental Setups}
\noindent {\bf Dataset and Evaluation Metric.} SNLI-VE dataset\cite{xie2019visual} and MS COCO\cite{chen2015microsoft} are used to evaluate our methods for visual entailment and image captioning. SNLI-VE dataset provides a pair of image and text and requires the model to determine if the given image and text description are correlated. We report accuracy and expected time reduction rate on dev and test datasets. MS COCO is used for evaluating the image captioning which requires models to generate an appropriate and fluent caption for a given image. The input of our model in image captioning is a image and a piece of instruction text. We report BLEU-4, METEOR, CIDEr, SPICE scores and expected time reduction rate on the Karpathy test split in main results. The full evaluation metrics can be found in supplementary details. As the measurement of runtime might not be stable even in the same environment, we propose a new metric to evaluate the efficiency in Seq2Seq architecture, called expected time reduction rate which can be defined as:
\begin{equation}
  1 - \cfrac{  \cfrac{n_E}{N_E} + \cfrac{\sum_{i=1}^{N_w} w_i \times n_D}{\sum_{i=1}^{N_w} w_i \times N_D} }{2},
\end{equation}
where $N_E$ and $N_D$ are the number of encoder and decoder layers in the model, $n_E$ and $n_D$ are the number of layers used in encoder and decoder during inference and $w_i$ is the number of words that exit at $n_D$ of decoder. The new metric is able to reflect the overall computation reduction ratio of Seq2Seq models.

\noindent {\bf Baselines.} To empirically evaluate the efficiency gains enabled by our proposed measurements, we compare with the original ${\bf OFA_{\mathit{Base}}}$\cite{wang2022ofa} model which includes 6 encoder layers and 6 decoder layers. ${\bf OFA_{\mathit{Tiny}}}$ is the tiny version of OFA which only contains 4 encoder and decoder layers. Since the experiments of {\bf DeeBERT}\cite{xin2020deebert} and {\bf PAEBB}\cite{zhou2020bpatience} were originally conducted on BERT\cite{devlin2018bert}, we implement it in decoder part on OFA architecture. We only compare our model with {\bf DeeCap}\cite{fei2022deecap} in image captioning.

\noindent {\bf Implementation Details.}
Our model is based on a unified vision language model called OFA \cite{wang2022ofa}. We utilize the Base model and released pre-trained weight to fine-tune on downstream tasks. There are 6 encoder layers and 6 decoder layers in the Base model. The hidden state size is 768. More hyperparameter settings can be found in supplementary details. All experiments are implemented by PyTorch and conducted on 4 RTX 6000 GPUs.



\subsection{Main Results}
\begin{table*}[t]
    \centering
    \begin{tabular}{l|cc|cccc}
    \hline
    \hline
        Task &Image Layer &Text Layer &BLEU-4 &METEOR &CIDERr &SPICE  \\
    \hline
        \multirow{4}*{Image Captioning}
        &6.0 &6.0 &42.4 &31.2 &143.9 &25.1 \\ 
        \cline{2-7}
        &3.1 &2.0 &32.8 &27.4 &112.1 &20.8\\
         &3.1 &6.0 &33.1 &27.4 &112.2 &20.7\\
         &6.0 &2.0 &\textbf{42.0} &\textbf{31.2} &\textbf{143.6} &\textbf{25.1}\\
           
    \hline
    \hline
    \end{tabular}
    \vspace{1em}

    \begin{tabular}{l|cc|cc}
    \hline
    \hline
        Task &Image Layer &Text Layer  &Dev &Test  \\
    \hline
        \multirow{4}*{Visual Entailment} 
        &6 &6  &88.6 & 88.7 \\
         \cline{2-5}
        &2.03 &2.9  &76.1 &75.6 \\
         &6 &2.9  &79.1 &79.5 \\
         &2.03 &6  &\textbf{88.4} &\textbf{88.6} \\

    \hline
    \hline
    \end{tabular}
    \caption{Results of our proposed decomposition strategy. The results show that for visual entailment, more image encoder layers can be safely removed and exited earlier while the text encoder layers for image captioning.}
    \label{tab:decomposition}
\end{table*}

\subsubsection{Evaluation on Visual Entailment}
We evaluate our methods on Visual Entailment, which is a classification task. Qualitative comparisons are shown in Table \ref{tab:result}. Our method has saved more computation and achieved best performance than any other method in the table. More concretely, our method can reduce nearly 50\% computation while only 0.7 points accuracy drop on the test set of SNLI-VE dataset. In comparison, $\rm OFA_{\mathit{Tiny}}$ model can save around 33\% computation with a significant performance drop, from 89.2 to 85.2. This is mainly because the low capacity of the tiny model limits the upper bound of performance. At the same time, DeeBERT is able to only reduce around 15\% of overall computation since DeeBERT cannot skip layers in encoder part, which requires large computation. Moreover, DeeBERT caused severe damage to accuracy, from 89.2 to 78.8. The reason behind this circumstance is that the shallow features are not sufficient for final prediction, as demonstrated in \cite{fei2022deecap}. Our proposed training objective can overcome the performance drop while no more computation is introduced. PABEE is better than DeeBERT in terms of accuracy with similar expected time reduction rate but still lower than our method. Overall, compared with previous early exiting methods, our model accelerates vision language model most with a slight accuracy reduction in visual entailment.
\subsubsection{Evaluation on Image Captioning}
\label{section:evalcaption}
We discuss the efficiency and effectiveness of our method in image captioning task. Experiment results are shown in Table \ref{tab:result}. Compared with other early exiting methods and tiny models, our model achieves the highest computation reduction, up to 40.2\%. Moreover, the performance of our method still ranks top among other methods on BLEU-4, METEOR, CIDEr and SPICE. DeeBERT causes the highest performance drop, showing the pure classifier is unable to identify the suitable time to exit. With performance recovery strategies, DeeCap can obtain better performance than DeeBERT and PABEE with more computation reduction. Our model appends cross-entropy loss on each layer of decoder which optimizes every decoder layer simultaneously. This training objective minimizes the error of a single layer and reduces the overall error accumulation in autoregressive decoding stages. MuE achieves higher performance on every metric comparing with DeeCap, which proves the effectiveness of our training objectives on performance recovery. 

\begin{figure*}[t]
\begin{center}
\includegraphics[width=0.313\textwidth]{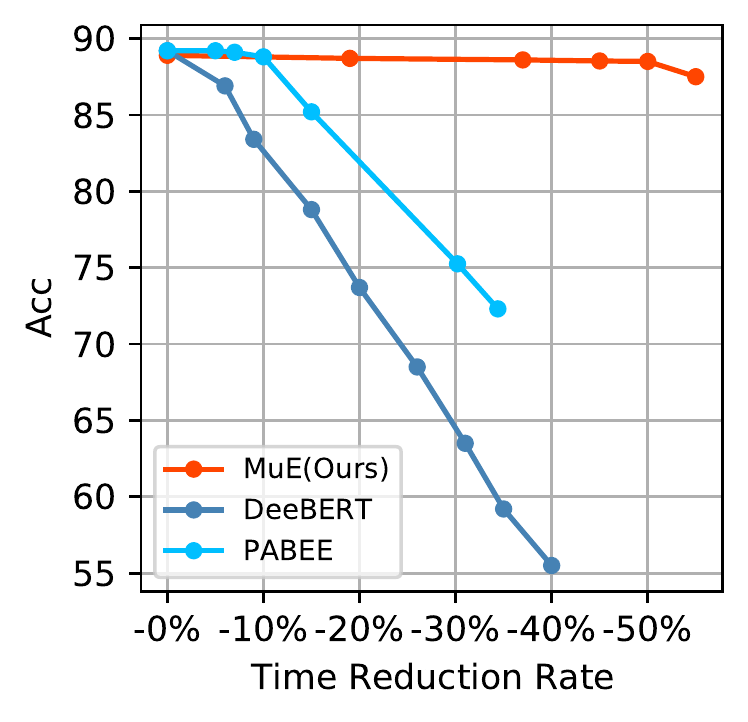}
\includegraphics[width=0.34\textwidth]{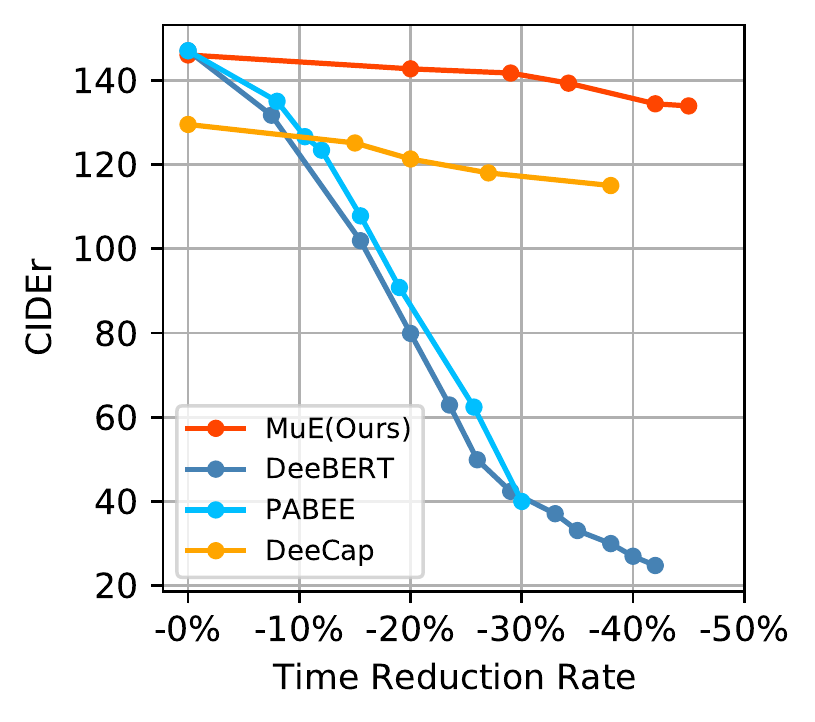}
\includegraphics[width=0.33\textwidth]{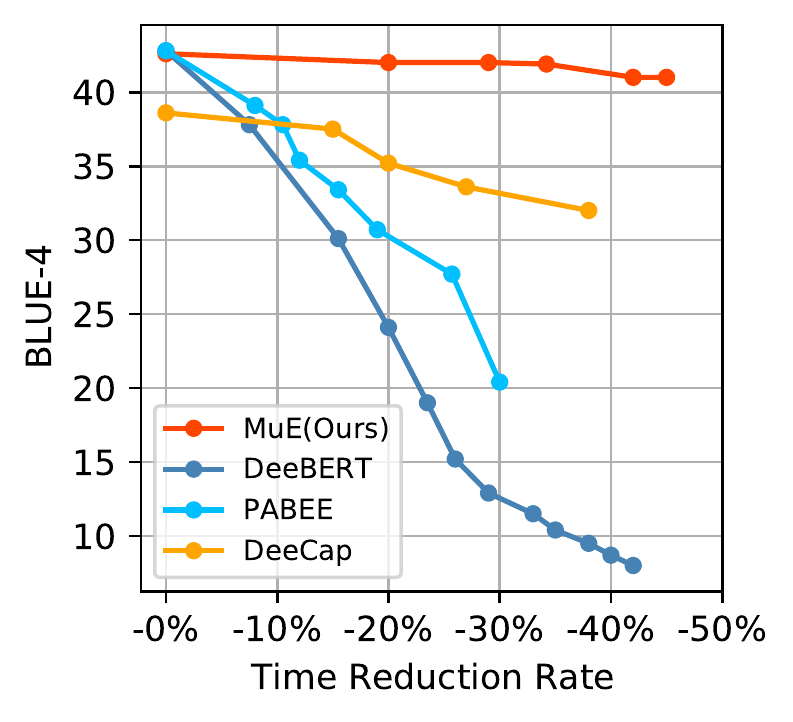}

\vspace{-2em}
\end{center}
\caption{Accuracy and expected time reduction rate trade-off comparison between our proposed MuE and other methods in visual entailment and image captioning.}
\label{Fig:tradeoff}
\end{figure*}

\vspace{-0.1in}

\subsubsection{Effectiveness of Decomposition}
We analyze our proposed modality decomposition strategy. In our experiments, we change the early exiting threshold to control the number of encoder layers that are utilized to process the tokens. We report the performance with different combinations of text and image encoder layers in visual entailment and image captioning. The main results are shown in Table \ref{tab:decomposition}. The last row is the model with full encoder layers for image and text. The second and third rows skip text and image separately. The first row skips both layers simultaneously. From the table, the accuracy drops dramatically when more text information is lost in visual entailment. The results show that for visual entailment, the understanding of text information is more important than image information. Conversely, in image captioning, all metrics decrease in varied degree if the image layers are skipped more. This demonstrates that the understanding of image information is more essential to image captioning. If more text layers and image layers are skipped at the same time, the performance is dropped in both tasks. Therefore, we should remove image and text layers selectively. Through decomposing image and text tokens in encoder, our model is able to exit earlier if the modality information is unnecessary while keeping more layers when the information is indispensable. For example,  the third row of visual entailment table shows that only 2.03 encoder layers that process image tokens is sufficient to obtain 99.8\% accuracy. This proves the effectiveness of our proposed decomposition strategy in the encoder.

\begin{table*}[t]
    \centering
    \begin{tabular*}{12.7cm}{l|ccc|ccccc}
    \hline
    \hline
        \multirow{2}*{Models} & \multicolumn{3}{c|}{SNLI-VE} & \multicolumn{5}{c}{Image Captioning} \\
        & Dev & Test & Time & BLEU-4 & METEOR & CIDEr & SPICE & Time \\
    \hline
       $\text{OFA}_{\textit{Base}}$ & 89.3 & 89.2 & 1 & 42.8 & 31.7 & 146.7 & 25.8 & 1 \\
    \hline
       \multirow{4}*{MuE} 
        & 88.8 & 88.7 & -19\% &42	&31.3	&142.7	&25.1 & -20.0\% \\
        & 88.7 & 88.6 & -37\% &42	&31.1	&141.7	&25.1 & -28.9\% \\
        & 88.6 & 88.5 & -45\% & 41.6 & 30.7 & 139.3 & 24.4 & -34.2\% \\
        & 87.7 & 87.5 & -55\% &41	&30	&133.9	&23.6 & -45\% \\
    \hline
    \hline
    \end{tabular*}
    \caption{Trade-off of our proposed MuE.}
    \label{tab:trade-off} \vspace{-0.1in}
\end{table*}


\begin{table}[ht]
    \centering 
\resizebox{1.0\linewidth}{!}{
    \begin{tabular}{l|c|c|c|c}
    \hline
    \hline
        Models &Test Acc. &Time &CIDEr &Time  \\
    \hline
         MuE Full &88.5 &-50\% &137.0 &-40.2\% \\
         MuE w/o Decom &75.0 &-35\% &117.5 & -21\%\\
         MuE w/o Task loss &87.0 &-47\% &117.9& -25\%\\
    \hline
    \hline
    \end{tabular}}
    \caption{Ablation study in visual entailment and image captioning. The results of our model without decomposition strategy and training objective. Time: time reduction rate}
    \label{tab:veablation}
\end{table}

\subsubsection{Accuracy and Computation Trade-off}
We analyze the trade-off between the accuracy and computation reduction of our model. According to Table \ref{tab:trade-off}, there are several experimental observations. First, the performance on visual entailment and image captioning drops with the increase of computation reduction. At the beginning, the large computation decrease doesn't do harm to performance. With the increase of skipped layers, little reduction in computation brings large amount of performance drop in some metrics such as CIDEr. This makes sense since if we force some mid-difficult samples to exit too early, they will be mistaken by output layers. Moreover, we draw the trade-off curves of our methods and other SoTA methods in Figure \ref{Fig:inspiration} and Figure \ref{Fig:tradeoff}. As shown in the figure, we can notice that comparing the trade-off curve of other methods, the curve of our method is generally located at the upper right of other curves. It means that with the same computation reduction rate, our method always obtains higher scores. Even at the highest
computation drop point, our method is much better than the performance of other methods such as DeeBERT and PABEE at the lowest computation reduction point. Interestingly, at the start point without computation reduction, the performance of our method is slightly lower than DeeBERT and PABEE. This is because we train our decomposition strategy with original fine-tuning and it brings slight degradation on overall performance.

\section{Ablation Study}

To prove whether our proposed methods are solid and how much these methods contribute to our model separately, we evaluated the model without decomposition strategy and training objective, respectively. The main results are shown in Table \ref{tab:veablation}. The first row is the full model with both decomposition and training strategies. We removed the decomposition strategy in the experiment at the second row. Therefore, in the encoder part, the image and text tokens early exit at the same layer while the decoder layer exiting strategy stays unchanged. In the experiment at the last row, we split training objectives from our full model to show the performance recovery brought by layer-wise task loss. All experiment results are the best trade-off between the scores and expected time reduction rate in visual entailment and image captioning.

As shown in the tables, the model without decomposition strategy gains the lowest test accuracy, CIDEr scores and expected time reduction rate. This is largely due to the information loss of key modality which is the text for visual entailment and image for image captioning. Coarse understanding in the description of the key modality is not enough to make a precise final prediction. Therefore, it is important to keep more layers for key modalities while remove more layers for other layers to improve efficiency. As for training objective, in visual entailment, removing training objective doesn't do harm to the final performance. However, the absence of our proposed layer-wise task loss causes a large drop in both performance and expected time reduction rate in image captioning. This is caused by the accumulation of errors in captioning while there is no accumulation in visual entailment with only one timestep in the decoding stage. Our layer-wise task loss is able to reduce error at every timestep of decoding which is beneficial to final results. The improvement demonstrates the strong ability of our proposed training objective to recover performance.

\section{Conclusion}
In this paper, we point out that different modality information is required for different tasks in unified vision language models. Moreover, previous early exiting methods cannot only be applied to the encoder part in Seq2Seq frameworks. To meet these features, we propose MuE for accelerating unified vision language models during the inference stage. We decompose the unified encoder into text and image encoders by splitting the input tokens of image text during fine-tuning stage. Similarity between layers is utilized as a metric to determine the exiting time which enables early exiting in both the encoder and decoder part in Seq2Seq architecture. In addition, in order to recover the performance drop caused by early exiting, we propose to train every decoder layer with the same training loss which helps to optimize every layer to the best performance. Experiments demonstrate that our proposed MuE reduces most computation with minimal performance drop in both classification and generation tasks, suggesting the effectiveness of our methods.


{\small
\bibliographystyle{ieee_fullname}
\bibliography{egbib}
}
\clearpage

\section*{Appendix}

\section*{A: More Experimental Setup}
We adopt OFA-base as the backbone for all experiments. OFA~\cite{wang2022ofa} is a sequence-to-sequence model which can unifies different modalities as well as tasks. During training, with a learning rate of 3e-5 and a batch size of 64, we train the model for 5 epochs on both visual entailment and image captioning tasks.  The embedding dimension is set as 768. We train the network using Adam optimizer with 0.01 weight decay.  For image captioning, the training has two stages, where the first stage is standard loss optimization and the second stage is specialized for CIDEr optimization. 

During inference, the similarity thresholds are set as 0.9 and 0.95 for the image and text encoder respectively. For visual entailment, the threshold for the decoder is set as 0.95. In image captioning, we utilize the threshold decay strategy as introduced in the method, where $\theta, \beta, \tau$ are set as 0.99, 0.95, 1, respectively.
\section*{B: Case Study of Image Captioning}
\begin{figure}[ht]
\begin{minipage}[c]{.2\textwidth} 
    \includegraphics[width=\textwidth]{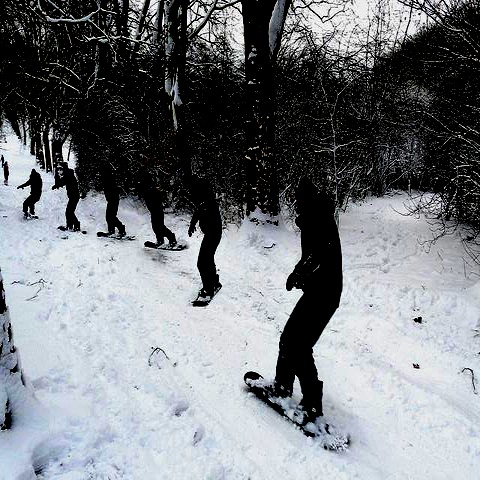} 
  \end{minipage} 
\begin{minipage}[c]{.3\textwidth} 
    \begin{flushleft}
        \textbf{GT}: six people are snow boarding down the hill.\\
    \textbf{DeeBERT}: a group of people in the same same thing in the snowboard.\\
    \textbf{PABEE}: a group of people are in the midst of a couple of people in  snow. \\
    \textbf{DeeCap}: a group of people are riding in the snow.\\
    \textbf{MuE}: a group of people riding snowboards down a snow covered slope.
    \end{flushleft}
    \label{fig:side:caption} 
  \end{minipage}%

\begin{minipage}[c]{.2\textwidth} 
    \centering 
    \includegraphics[width=\textwidth]{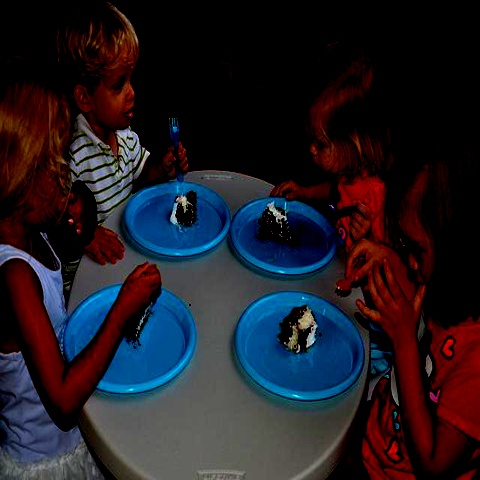} 
  \end{minipage} 
\begin{minipage}[c]{.3\textwidth}
    \begin{flushleft} 
    \textbf{GT}: a group of children sitting at a table eating pieces of cake.\\
    \textbf{DeeBERT}: a group of people sitting around a table with a bunch of food.\\
    \textbf{PABEE}: a group of people who are sitting at a table with bunch of food.\\
     \textbf{DeeCap}: a group of children who are sitting at a table with bunch of food.\\
    \textbf{MuE}: a group of children sitting at a table eating a slice of cake.
    \end{flushleft}
  \end{minipage}%

\caption{Case Studies on image captioning.}
\label{Fig:case}
\end{figure}
Following previous work~\cite{fei2022deecap} and in order to have more intuitive understanding, we provide several examples of captions generated by different methods and corresponding ground truth. As shown in Figure~\ref{Fig:case}, all captions generated by different models can represent image meanings accurately. At the same time, our MuE model can generate captions with semantic meanings. For example, MuE can realize detailed information in images such as "children" and "cake" compared with "people" and "food". This demonstrates the effectiveness of our model in reducing generation errors.
\section*{C: Experimental Results in Image Captioning}

\begin{figure*}[ht]
\begin{center}
\includegraphics[width=0.33\textwidth]{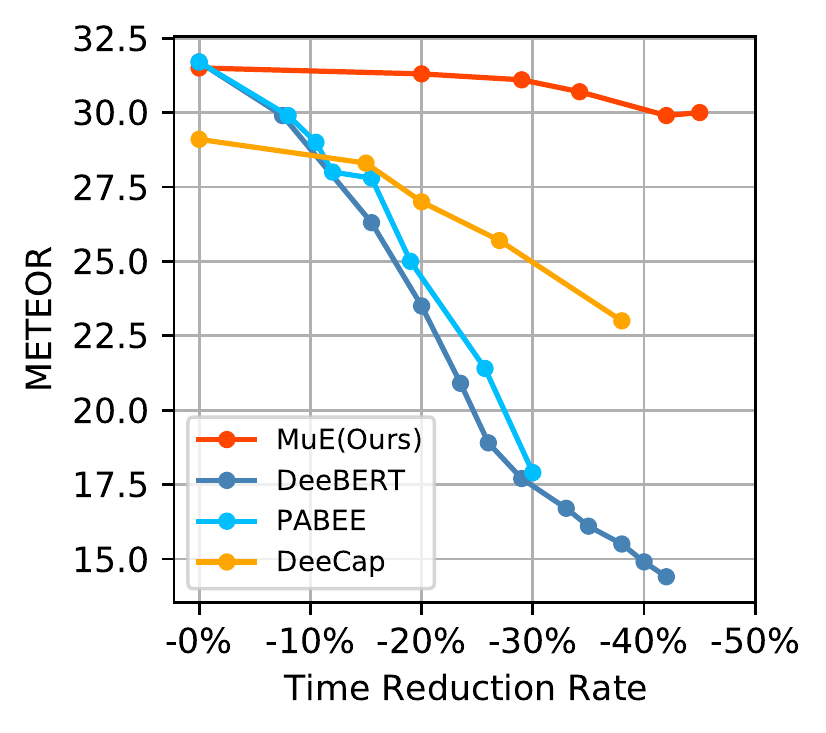}
\includegraphics[width=0.33\textwidth]{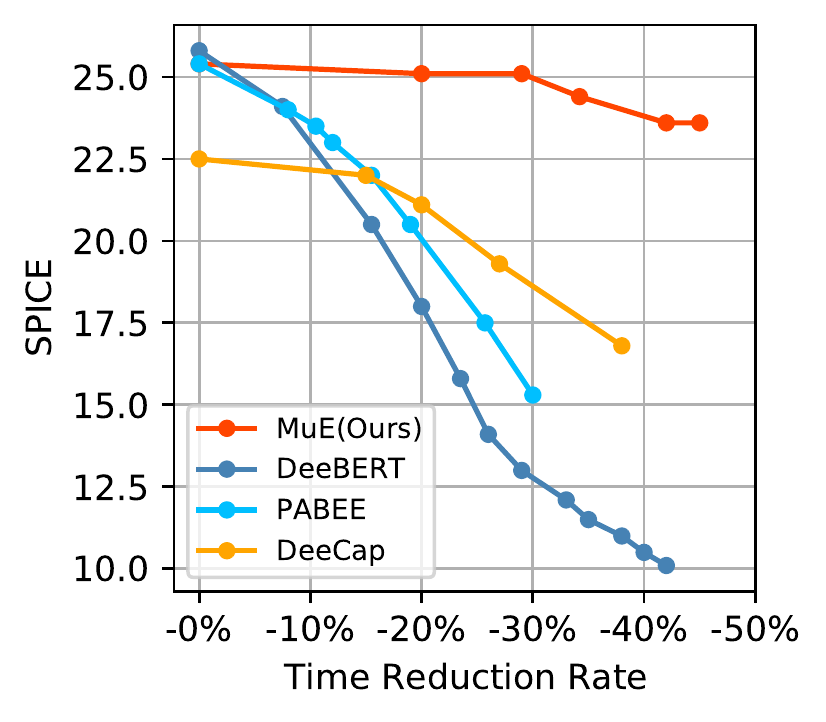}
\includegraphics[width=0.315\textwidth]{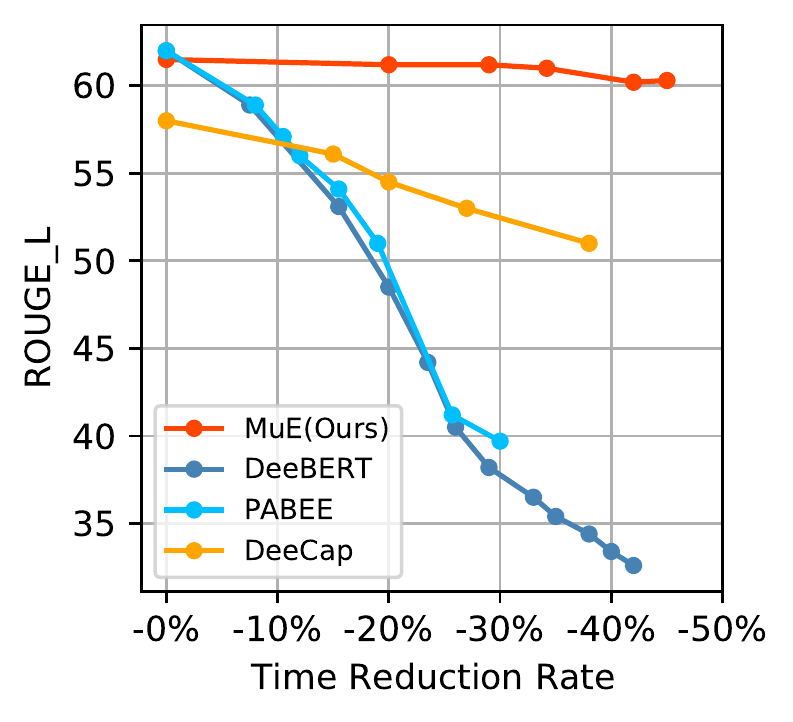}

\vspace{-2em}
\end{center}
\caption{Task performance changes with respect to varying time reduction rates. }
\label{Fig:add_tradeoff}
\end{figure*}

{The more detailed results on image captioning are included in Table~\ref{tab:add_result}  and Figure~\ref{Fig:add_tradeoff}. Table~\ref{tab:add_result} shows the performance comparison in term of several metrics and corresponding expected time reduction percentage. We observe that the proposed method is able to achieve best performance in term of various metrics and achieve largest time reduction on image captioning. Figure~\ref{Fig:add_tradeoff} shows the task performance changes with respect to varying time reduction rates, illustrating that the proposed method MuE is able to maintain most of the performance as the time reduction rate increases comparing to other methods. More specifically, the proposed method can reduce nearly 50\% computation with minimal performance drop while other methods suffer from significant performance drop.}


\begin{table}[h]
    \centering
    \resizebox{1.0\linewidth}{!}{
    \begin{tabular*}{10cm}{l|cccccc}
    \hline
    \hline
        \multirow{2}*{Models} & \multicolumn{5}{c}{Image Captioning} \\
         & BLEU-1 & BLEU-2 & BLEU-3 & ROUGE-L & Time \\
    \hline
       $\text{OFA}_{\textit{Base}}$  & 83.4 & 68.3 & 54.1 & 62.1 & 1 \\
    \hline
       $\text{OFA}_{\textit{Tiny}}$  & 74.3 & 60.8 & 48.2 & 55.3 & -33\% \\
       DeeBERT  & 68.1 & 46.1 & 33.7 & 53.1 & -15.5\% \\
       PABEE  & 70.3 & 55.8 & 42.3 & 54.1 & -16.3\% \\
       DeeCap &75.5 &62.0 & 50.1 & 57.2 & -38\% \\
    \hline
       Ours  & \textbf{81.2} & \textbf{66.8} & \textbf{52.9} & \textbf{60.2} & \textbf{-40.2\%} \\
    \hline
    \hline
    \end{tabular*}}
    \caption{The performance and expected time reduction rate comparison in term of various image captioning evaluation metrics. Our method reduces largest computation comparing to other methods while preserving the most of performance. We use ``Time'' as a short for expected time reduction rate.}
    \label{tab:add_result}
\end{table}

\section*{D: More Analysis of MuE}

\noindent{\bf Low cost to decide early exiting.} 
{The existing early exiting methods usually adopt linear classifiers to simulate exiting performance in each layer.  However, the computation of adding classifiers for early exiting purpose is non-negligible for a large model with multiple layers. Such a case become worse for generation tasks, where the classifier needs to calculate probabilities among a large vocabulary size. Different with existing methods, the proposed methods depend on layer-wise similarities and reduce the probability calculation over a large dimension,  resulting in computation cost reduction.}


\noindent{\bf Challenge when 50\% computation reduction is required.}
{As discussed in the limitation section, we observe performance drop of our method on image captioning when 50\% computation reduction is required. Even though the proposed method still outperforms other methods with the same computation reduction requirement, we are interested in why the performance is difficult to be maintained and discuss potential causes and solutions.  We hypothesis this may be attributed to task characteristics, where the image captioning task may need more knowledge or layers than visual entailment to maintain the performance. Towards this, we propose to skip the image tokens instead of layers for this task. The motivation is from an observation that image tokens are somewhat redundancy. But we plan to leave this problem for future works.}

\end{document}